\documentclass[11pt]{article}

\usepackage{acl}

\usepackage{times}
\usepackage{latexsym}

\usepackage[T1]{fontenc}

\usepackage[utf8]{inputenc}

\usepackage{microtype}

\usepackage{inconsolata}

\usepackage{graphicx}
\usepackage{blindtext}
\usepackage{amsfonts}
\usepackage{amsmath}
\usepackage{booktabs}

\title{
\raisebox{-0.3\height}{\includegraphics[width=1cm]{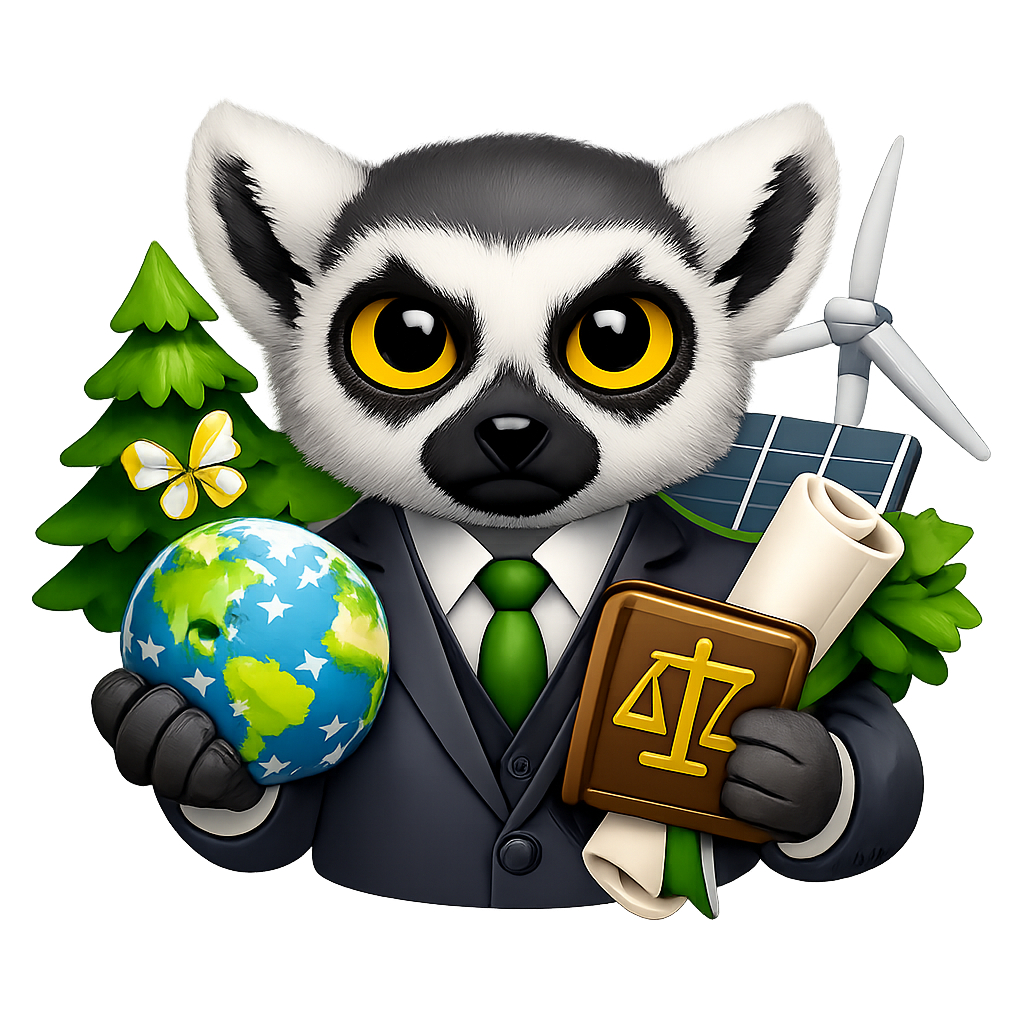}} 
LEMUR: A Corpus for Robust Fine-Tuning of \\Multilingual Law Embedding Models for Retrieval
}

\author{
 \textbf{Narges Baba Ahmadi},
 \textbf{Jan Strich},
 \textbf{Martin Semmann,}
 \textbf{Chris Biemann}\\
 Hub of Computing and Data Science (HCDS) \\
 University of Hamburg, Germany \\
 \\
 \small{
   \textbf{Correspondence: \texttt{\{first\_name\}.\{last\_name\}@uni-hamburg.de}}
 }
}

\begin{document}
\maketitle
\begin{abstract}

Large language models (LLMs) are increasingly used to access legal information. 
Yet, their deployment in multilingual legal settings is constrained by unreliable retrieval and the lack of domain-adapted, open-embedding models. 
In particular, existing multilingual legal corpora are not designed for semantic retrieval, and PDF-based legislative sources introduce substantial noise due to imperfect text extraction. 
To address these challenges, we introduce LEMUR, a large-scale multilingual corpus of EU environmental legislation constructed from 24,953 official EUR-Lex PDF documents covering 25 languages. 
We quantify the fidelity of PDF-to-text conversion by measuring lexical consistency against authoritative HTML versions using the Lexical Content Score (LCS).
Building on LEMUR, we fine-tune three state-of-the-art multilingual embedding models using contrastive objectives in both monolingual and bilingual settings, reflecting realistic legal-retrieval scenarios. 
Experiments across low- and high-resource languages demonstrate that legal-domain fine-tuning consistently improves Top-k retrieval accuracy relative to strong baselines, with particularly pronounced gains for low-resource languages. 
Cross-lingual evaluations show that these improvements transfer to unseen languages, indicating that fine-tuning primarily enhances language-independent, content-level legal representations rather than language-specific cues. 
We publish code\footnote{\href{https://github.com/nargesbh/eur_lex}{GitHub Repository}} and data\footnote{\href{https://huggingface.co/datasets/G4KMU/LEMUR}{Hugging Face Dataset}}.

\end{abstract}

\section{Introduction}

LLMs are transforming legal work and research by making access to legal knowledge, automated document review, and case law summarization significantly faster and easier~\cite{zheng_when_2021}.
However, the deployment of these models in legal practice is often hindered by "hallucinations" and a lack of grounding in authoritative legal sources \cite{reuter_towards_2025, magesh_hallucination-free_2025}. To mitigate these risks, Retrieval-Augmented Generation (RAG) has become the de facto standard architecture, ensuring that model outputs are anchored in verifiable primary documents~\cite{lewis_retrieval-augmented_2020}.

While RAG relies on the combination of an LLM for generation and embedding models for retrieval, its success is fundamentally dependent on the retrieval setup and the embedding model used for the vector database \cite{gao_retrieval-augmented_2023}. 
While these models are typically general-purpose, fine-tuning them on domain-specific data consistently yields superior performance compared to existing specialized models~\cite{tang_we_2025}, particularly in law, where text often contains archaic terminology, complex syntactic structures, or polysemy~\cite{ariai_natural_2025}.
Nevertheless, these models are untrained in the legal domain and primarily monolingual~\cite{chalkidis_legal-bert_2020, chalkidis_lexglue_2022} or proprietary~\cite{voyage_ai_voyage_2024}.

While multilingual datasets in law already exist, the data are typically formatted for pretraining~\cite{henderson_pile_2022, niklaus_multilegalpile_2024} or for classification of legal documents~\cite{chalkidis_multieurlex_2021}, leaving a void for high-quality benchmarks dedicated to cross-lingual semantic retrieval. Furthermore, legal corpora are often stored in PDF format, which can introduce inaccuracies when converted to text for effective search due to multi-column layouts and nested tables. This 'extraction gap' affects data integrity in RAG systems, as downstream embedding models are forced to process corrupted or misaligned tokens.
To address these gaps, our contributions are:
\begin{itemize}
    \item \textbf{Multilingual Dataset (LEMUR):} We introduce a \textbf{L}aw \textbf{E}uropean \textbf{MU}ltilingual \textbf{R}etrieval corpus (\textbf{LEMUR}), which consists of \textbf{25k} EU legal PDFs in 25 languages, designed for training embedding models on legal text. \\
    \item \textbf{The Lexical Content Score (LCS):} We systematically analyze PDF-to-text conversion quality by measuring content consistency across \textbf{twenty-five} languages.
    \item \textbf{Legal Embedding Fine-Tuning:} We fine-tune \textbf{three SOTA} embedding models on \textbf{five languages} for a legal document retrieval task and evaluate them in monolingual, bilingual, and cross-lingual settings.
\end{itemize}

\section{Related Work}

\paragraph{Multilingual Legal Corpora.}

Research on multilingual legal corpora has produced both supervised benchmarks \citep{chalkidis_legal-bert_2020, chalkidis_lexglue_2022, zheng_when_2021, ma_lecard_2021} and large-scale pretraining resources \citep{niklaus_multilegalpile_2024, el-haj_multilingual_2024}. \citet{chalkidis_multieurlex_2021} introduces \textsc{MultiEURLEX}, a multilingual multi-label dataset of EU legislation in 23 languages for legal document classification, while \citet{chalkidis_lexglue_2022} proposes \textsc{LexGLUE}, a suite of English legal NLU benchmarks that has become a standard evaluation protocol for legal language models. Beyond EU legislation, \citet{zheng_when_2021} presents CaseHOLD, a multiple-choice benchmark comprising more than 53{,}000 U.S. case-law holdings, and \citet{ma_lecard_2021} introduces LeCaRD, a large-scale case-retrieval dataset for the Chinese criminal law system with expert-designed relevance criteria.

Our work contributes to this line of research by constructing a new multilingual EU law dataset directly from official legislative PDFs and targeting downstream embedding-model fine-tuning across multiple European languages, thereby bridging large-scale pretraining corpora and task-specific benchmarks in an EU legislative setting.

\paragraph{Embedding Models and Legal Retrieval.}

Recent work shows that structure-aware models such as \textsc{SAILER}~\citep{li_sailer_2023} and \textsc{DELTA}~\citep{li_delta_2025} capture section-level or structural dependencies to improve legal case retrieval, while \textsc{SM-BERT-CR}~\citep{vuong_sm-bert-cr_2022} and \textsc{ReaKase-8B}~\citep{tang_reakase-8b_2025} incorporate supporting-relation modeling and reasoning-driven representations. For multilingual and cross-lingual settings, \textsc{LexCLiPR}~\citep{upadhya_lexclipr_2025} enables paragraph-level retrieval across ECtHR judgments, showing that off-the-shelf multilingual encoders struggle without domain-adaptive training.

Domain-specific pretraining consistently improves legal NLP tasks. \citet{limsopatham_effectively_2021} shows that in-domain pretraining and long-document handling benefit legal classification, while \citet{darji_german_2023} demonstrates gains from adapting BERT to legal NER over BiLSTM–CRF baselines. More broadly, \citet{tang_we_2025} and BloombergGPT \citep{wu_bloomberggpt_2023} provide evidence that domain-adapted embeddings remain essential despite strong general-purpose LLMs.

Although these studies focused mainly on monolingual legal data or non-legal domains, they have not systematically studied cross-lingual retrieval for EU legislation. We address this gap by introducing a multilingual EU law corpus and evaluating fine-tuned embedding models for monolingual and cross-language retrieval on EUR-Lex texts.

\section{LEMUR}\label{lemur}

We construct \textsc{LEMUR} from official documents published on EUR-Lex\footnote{\url{https://eur-lex.europa.eu/homepage.html}}.
Section~\ref{sec:document-collection} details how the source documents are identified, selected, and collected from the EUR-Lex repository. 
Section~\ref{sec:pdf-to-text} then explains the process of converting the original PDF files into a structured and machine-readable text format, 
and Section~\ref{sec:data-preprocessing} describes the subsequent preprocessing steps, including the construction of high-quality query--document pairs used in our experiments.
An overview of the data preparation process from EUR-Lex PDFs to structured JSONL is shown in Figure~\ref{fig:end-end-pipeline}.

\subsection{Document Collection}\label{sec:document-collection}

\begin{figure}[t]
    \centering
    \includegraphics[width=0.95\linewidth]{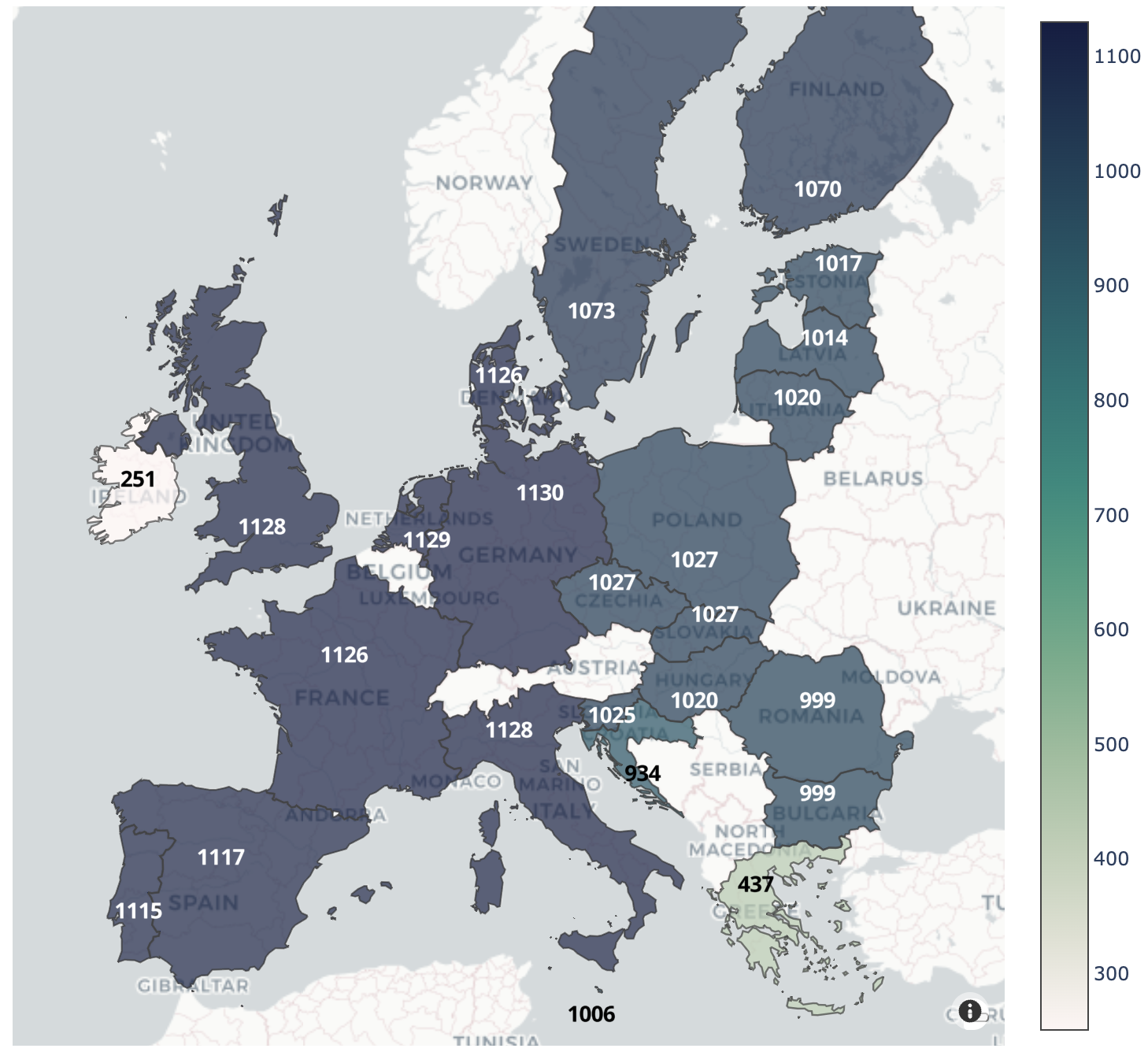}
    \caption{Number of documents per country in LEMUR.}
    \label{fig:map_plot}
\end{figure}

To build a focused corpus, we gathered all legal acts listed under \emph{Category~15} (Environment, consumers and health protection), \emph{Subcategory~10} (Environment), across all available publication years. This yielded \textbf{1{,}174} distinct legal acts from \textbf{1961–2025}.  
Because each act is available in 25 official EU languages, the collection comprises a total of \textbf{24{,}953} PDF documents and results in \textbf{~461k} pages. Figure~\ref{fig:map_plot} summarizes the number of records per language.
Coverage is highest for languages with long-standing EU membership (e.g., German, Dutch, English, Italian), and lower for countries with more recent membership (e.g., Croatia or Bulgaria).

\begin{figure}[t]
    \centering
    \includegraphics[width=\linewidth]{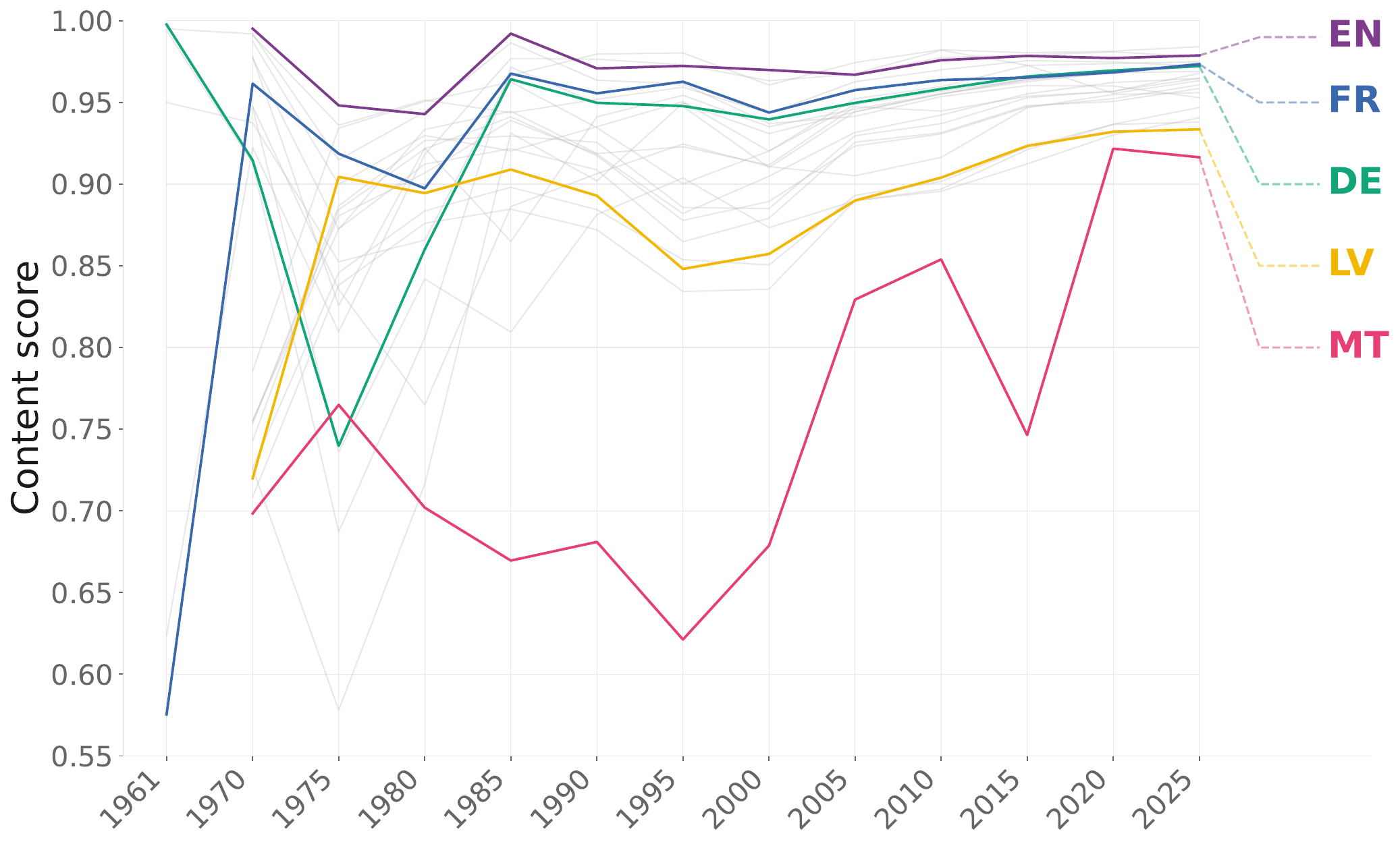}
    \caption{Average Content Score similarity per year (5-year bins) for the five languages used in our experiments}
    \label{fig:Content Score-lang-year}
\end{figure}

\subsection{PDF--to--Text Conversion}\label{sec:pdf-to-text}
In the original source of EUR-Lex the dataset is available in PDF and HTML format. Previous datasets~\cite{chalkidis_multieurlex_2021} used the HTML version, but we found that tables were not converted correctly.
Therefore, we tested multiple PDF--to--text services (Docling~\cite{livathinos_docling_2025}, Unstructured~\cite{unstructured_team_unstructured_2023}, PyMuPDF~\cite{pymupdf_developers_pymupdf_2021}) but found that the best results were obtained by converting all PDFs into structured JSONL files using \textbf{olmOCR}~\citep{poznanski_olmocr_2025}. On average, documents contain approximately \textbf{19} pages, with approximately \textbf{403} tokens per page, yielding roughly \textbf{7{,}781} tokens per document. These values indicate that LEMUR consists primarily of long-form legislative text, making it well-suited for evaluating embedding models on long-document and multilingual retrieval tasks.

To verify the quality of the PDF--to--text conversion, we compare each converted document against the corresponding HTML version available on EUR-Lex. While HTML files provide a clean textual baseline, they often linearise tables in ways that differ from the official PDF layout. In contrast, the JSONL files extracted with \textsc{olmOCR} preserve table structure more consistently and also in markdown format, which is essential for downstream retrieval tasks that rely on faithful representation of legislative formatting. For this reason, the JSONL representation is used as the primary data source in LEMUR, while the HTML version serves solely as a reference for evaluation. We present LCS for all approaches in Appendix~\ref{apx:content-score-methods}.

\paragraph{Lexical Content Similarity (LCS).}

To evaluate the PDF-to-text conversion, we compute a content similarity score between each converted document and its corresponding HTML version. Before that, the HTML text is normalized to remove superficial differences that could affect lexical comparison. This includes removing all styling attributes (e.g., class, id, style) from HTML tags, stripping leading and trailing whitespace, converting to lowercase, normalizing numeric formatting (e.g., \$ 100 becomes \$100), and collapsing repeated punctuation (e.g., ... is replaced with .).

After normalization, we represent both as bag-of-words vectors~\cite{qader_overview_2019}, $\mathbf{v}\text{H}$ and $\mathbf{v}\text{PDF}$, in a shared vocabulary of size $n$, where each entry corresponds to the frequency of a unique word. The content similarity score is then defined as the cosine similarity between these vectors, as

\begin{equation}
    \text{LCS}(h_\text{H}, h_\text{PDF}) = \frac{\sum_i v_{\text{H},i} \cdot v_{\text{PDF},i}}{\sqrt{\sum_i v_{\text{H},i}^2} \; \sqrt{\sum_i v_{\text{PDF},i}^2}}
\end{equation}

where $v_{\text{H},i}$ and $v_{\text{PDF},i}$ denote the counts of the $i$-th word in the HTML and PDF texts, respectively. By applying these preprocessing steps and computing cosine similarity, the content score  measures the actual lexical similarity between documents.

\begin{figure*}[t]
  \centering
  \makebox[\textwidth][c]{%
    \includegraphics[width=1\textwidth]{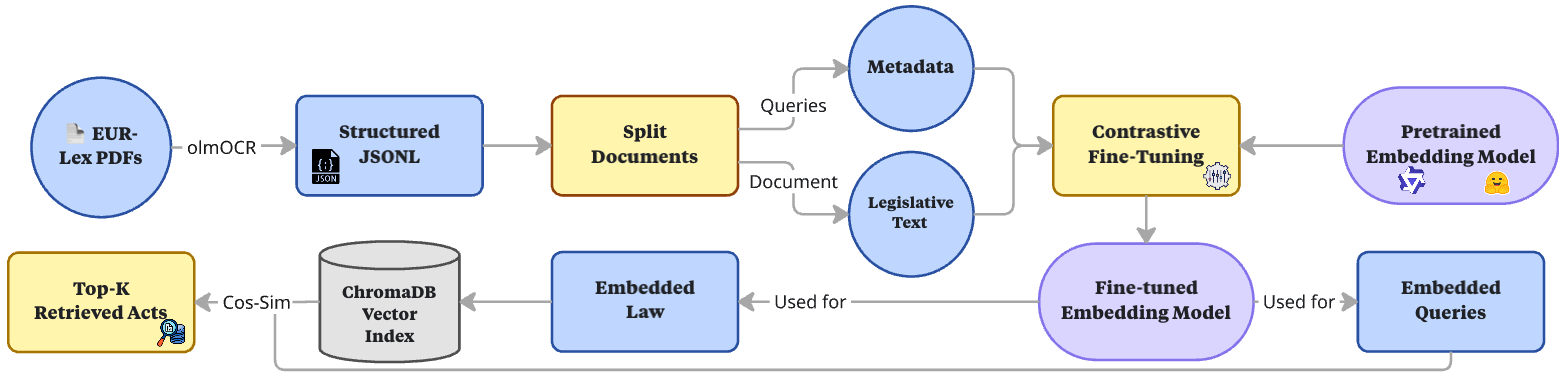}%
  }
  \caption{End-to-end pipeline for data preparation, contrastive fine-tuning, and retrieval. EUR-Lex PDFs are processed into structured JSONL, split into queries (metadata) and documents (legislative text), and used to fine-tune embedding models. The resulting embeddings are indexed for Top-$k$ retrieval of legislative acts.}
  \label{fig:end-end-pipeline}
\end{figure*}

\paragraph{Conversion Results.}
Figure~\ref{fig:Content Score-lang-year} illustrates the average Content Score similarity between the converted JSONL documents and their original HTML counterparts across all languages in LEMUR, stratified by year. For our primary analysis, we evaluate five languages with varying degrees of representation: high-resource languages (English (EN), German (DE), and French (FR)) and low-resource languages (Latvian (LV) and Maltese (MT)). This selection allows us to assess whether the model's performance generalizes from well-represented languages in the pretraining corpus to those that are comparatively underrepresented.

Our results indicate that for high-resource languages, the OLM-OCR model~\cite{poznanski_olmocr_2025} achieves a similarity score exceeding 95\%. However, we observe performance degradation for older documents, likely due to less standardized formatting compared with modern web documents. We hypothesize that the OLM-OCR training distribution is more closely aligned with contemporary layout standards. 
While performance is lower for low-resource languages, averaging approximately 90\% for Latvian and 80\% for Maltese, the similarity scores remain sufficiently high to justify using these converted documents for fine-tuning embedding models.
We also present avg. LCS for all other languages in Appendix~\ref{sec:avg_content_score} as well as an overview of the distribution of publications per year in Appendix~\ref{apx:content-score-by-year}. 

\subsection{Data Preprocessing}\label{sec:data-preprocessing}
Figure~\ref{fig:end-end-pipeline} shows the pipeline for the data preprocessing to transform the documents to query--document pairs. After the transformation to structured JSONL, each legislative document in LEMUR begins with a short introductory block that we refer to as \emph{metadata}. This block typically includes the act type (e.g., Commission Decision), the date, a brief description of the subject matter, references to the underlying legal basis, and standard publication notes, such as notification numbers, statements regarding the authentic language version, and indications of whether the text is relevant to the EEA.

We split each document into two parts: the introductory metadata block at the beginning of the document, which serves as the query, and the remaining substantive text of the legal act, which constitutes the document to be retrieved.
This setup reflects realistic legal search behavior: a user begins with a short, structured description that provides only partial information about the act, whereas the retriever must identify the full legislative text.
We use metadata as queries and the remaining text as a corpus, producing a large set of retrieval-ready pairs for both monolingual and cross-lingual evaluation. Examples are provided in Appendix~\ref{apx:dataset-examples}.

\section{Method}
This section gives an overview of the training procedure of the embedding models on LEMUR and the construction of our retrieval pipeline. Subsection~\ref{sec:data-pairs} outlines the retrieval-oriented data representation, while Subsection~\ref{sec:mono-finetune} presents our monolingual fine-tuning procedure based on a contrastive learning approach~\citep{hadsell_dimensionality_2006, henderson_efficient_2017}. Subsection~\ref{sec:bi-finetune} extends this approach to bilingual multi-positive training. Subsection~\ref{sec:rag} details the construction of the vector-database component used for retrieval.

\subsection{Retrieval-Oriented Training Pairs}
\label{sec:data-pairs}

As described in Section~\ref{lemur}, every document in LEMUR is split into a short metadata block and the remaining substantive legislative text. We directly adopt that structure for retrieval.

Accordingly, each legislative act yields a single query--document pair without requiring additional query construction or rewriting. This setup reflects realistic legal search behavior, in which users often begin with brief structured information. 
Each data entry contains the complete page content, with a clear separation between the metadata block and the remainder of the legislative text.
The data is split into \textbf{60\% training}, \textbf{20\% validation}, and \textbf{20\% test} sets, independently for each language or language pair, such that the same underlying legislative acts are assigned to the same split across languages, with each split containing the corresponding translations of those acts.

\subsection{Monolingual Contrastive Fine-Tuning}
\label{sec:mono-finetune}

We first adapt embedding models to the EUR-Lex retrieval setting in a \textbf{monolingual} fashion, fine-tuning one model per language. We experiment with the publicly available embedding models \texttt{Qwen3-0.6B} and \texttt{Qwen3-4B}~\citep{yang_qwen3_2025}, as well as \texttt{E5-Multilingual}~\citep{wang_multilingual_2024}, all obtained from the \textsc{MTEB} leaderboard\footnote{\url{https://huggingface.co/spaces/mteb}}.
These models were selected to cover a range of sizes and to have been pretrained on multilingual data and legal-domain tasks. They are also widely used in production and scored high on the MTEB leaderboard (3M downloads per month, Dec 25)\footnote{\url{https://huggingface.co/intfloat/multilingual-e5-large}}.
For each model and language, a dedicated embedding model is fine-tuned using metadata as queries and the corresponding legislative text as the positive document.
Fine-tuning uses a contrastive \emph{Multiple Negatives Ranking} (MNR) objective with in-batch negatives \citep{henderson_efficient_2017}. 

\paragraph{Objective Function.}
Given a batch of query--document pairs $\{(q_i, d_i)\}_{i=1}^{B}$, each $(q_i,d_i)$ is treated as a positive pair, while all other documents in the batch act as negatives. Let $f(\cdot)$ denote the encoder producing $L_2$-normalized embeddings, and let $s_{ij} = f(q_i)^\top f(d_j) / T$ denote the temperature-scaled cosine similarity.
We optimize the symmetric MNR loss:
\begin{equation}
\mathcal{L}
= -\frac{1}{2B} \sum_{i=1}^{B}
\left(
\log \frac{e^{s_{ii}}}{\sum_{j} e^{s_{ij}}}
+
\log \frac{e^{s_{ii}}}{\sum_{j} e^{s_{ji}}}
\right)
\end{equation}

\paragraph{Training Setup.}
We train for up to 30 epochs, with early stopping based on the validation loss. Most models support a maximum sequence length of 2{,}048 tokens; the only exception is \texttt{E5-Multilingual}, which is restricted to 512 tokens. Training uses \texttt{bfloat16} precision, gradient checkpointing where supported, and a linear warm-up schedule.
Training is performed on NVIDIA RTX A6000 and NVIDIA A100 (80GB) GPUs, with the larger \texttt{Qwen3-4B} model trained on A100 due to its higher memory requirements. In terms of training cost, fine-tuning \texttt{E5} typically completes within approximately 20--30 minutes per language, \texttt{Qwen3-0.6B} requires on the order of 2--4 hours, and \texttt{Qwen3-4B} requires roughly 6--8 hours per language, depending on the dataset size.

\begin{figure*}[t]
  \centering
  \makebox[\textwidth][c]{%
    \includegraphics[width=1\textwidth]{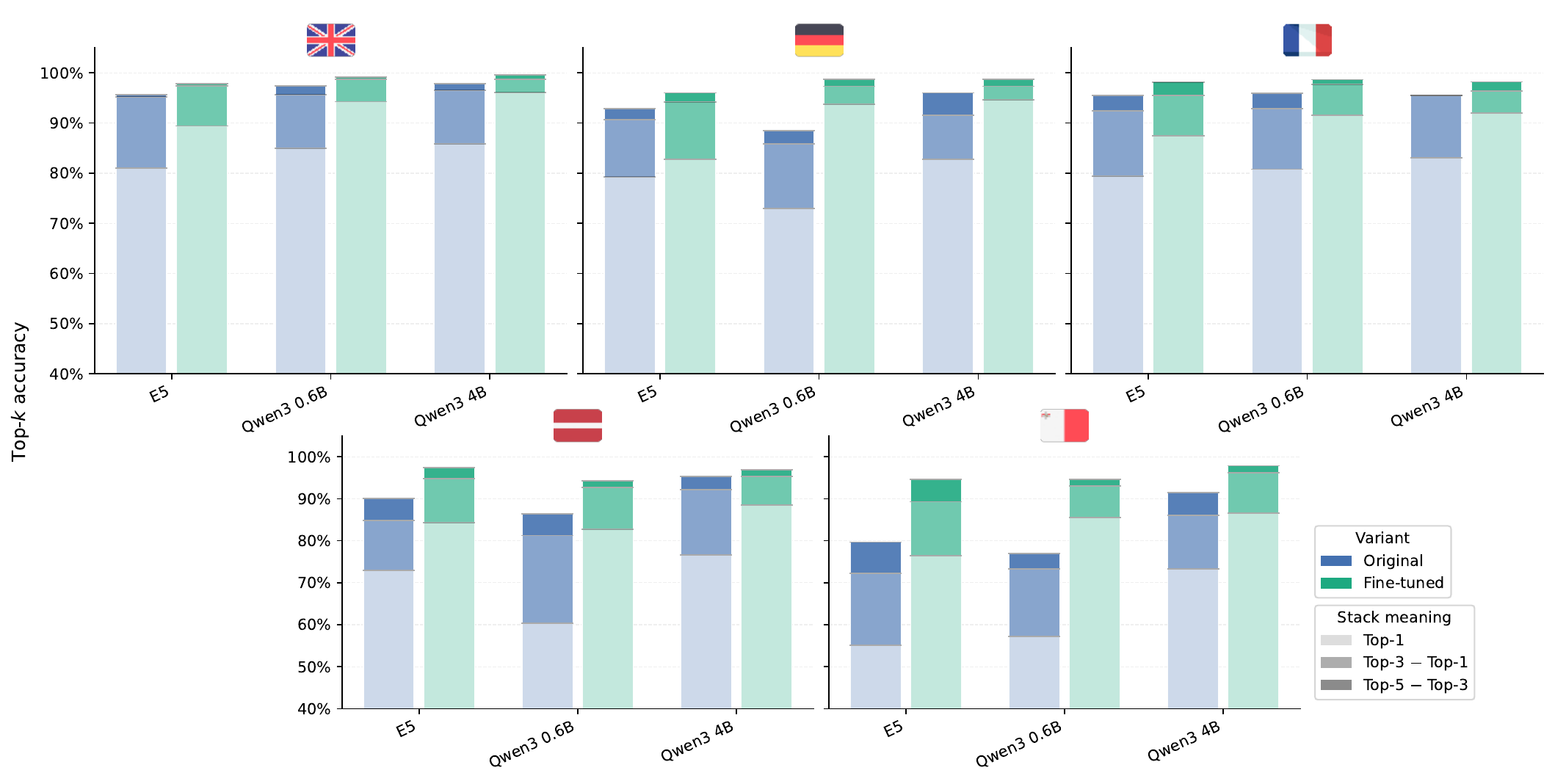}%
  }
  \caption{Monolingual fine-tuning of three embedding models (E5, Qwen-0.6B \& Qwen-4B) on five languages (EN, DE, FR, LV, MT). Performance is measured using Acc@\textit{k} for 1/3/5  on test queries evaluated against the test document collection, represented as stacked bars, and compared between the base model and the fine-tuned variant.}

  \label{fig:test-ds-test-search}
\end{figure*}

\subsection{Bilingual Multi-Positive Contrastive Fine-Tuning}
\label{sec:bi-finetune}

To exploit the availability of parallel legislative acts across languages, we extend the monolingual setup to a \textbf{bilingual multi-positive} scenario. In this setting, one metadata query is paired with \emph{multiple} language versions of the same legislative act, and all corresponding documents are treated as positives during training. This enables the model to learn jointly from aligned legal content across languages.

\paragraph{Objective Function.}
We use a \textbf{grouped multi-positive} extension of the symmetric MNR loss, following~\citet{zhao_leveraging_2024}. For each query embedding $q_i$, all document embeddings corresponding to aligned versions of the same legislative act are treated as positives, while all other documents in the batch serve as negatives. This objective encourages each query to be simultaneously close to multiple positive documents, promoting cross-lingual semantic alignment.

Similarity is computed using $L_2$-normalized embeddings and a temperature-scaled dot product. We optimize the following symmetric grouped multi-positive MNR objective:
\begin{equation}
\begin{aligned}
\mathcal{L}
= -\frac{1}{2B} \sum_{i=1}^{B} \Bigg[
&\log \frac{\sum_{j \in \mathcal{P}(i)} e^{s_{ij}}}
           {\sum_{j} e^{s_{ij}}}
\\
&+ \log \frac{e^{s_{ii}}}
              {\sum_{j} e^{s_{ji}}}
\Bigg]
\end{aligned}
\end{equation}
where $\mathcal{P}(i)$ denotes the set of positive documents for query $q_i$ within the batch.

\subsection{VectorDB Construction for Retrieval}
\label{sec:rag}

To test the performance of the embedding models, we simulated a retrieval by constructing a vector store using ChromaDB~\cite{chroma_team_chroma_2025}, a lightweight vector database optimized for similarity search.
For each language, we created a collection for both the base and fine-tuned embedding models.
Very long documents are truncated using a sequence of decreasing token caps to ensure compatibility with model limits. Across languages and models, approximately 8--15\% of documents require truncation; for these documents, roughly 40--50\% of their original tokens are removed. All stored vectors are $L_2$-normalized, and cosine similarity is used during retrieval.

At inference time, the metadata again serves as the query. It is embedded using the same model that indexed the documents, and nearest-neighbor search is performed in ChromaDB using cosine similarity to retrieve the most semantically similar legislative texts. This retrieval component constitutes the retrieval pipeline used in our experiments.

\section{Evaluation and Results}
\label{sec:eval-results}

This section outlines the three main experiments we conducted to demonstrate that multilingual embedding models can be trained on law data.
All experiments follow the same pipeline shown in Figure~\ref{fig:end-end-pipeline}, differing only in the fine-tuning configuration and language setup.
Firstly, we used monolingual contrastive fine-tuning to train on five individual languages (EN, DE, FR, LV \& MT) as described in Subsection~\ref{sec:mono-results}. Secondly, we conducted a bilingual fine-tuning experiment to study the interaction between high-resource and low-resource languages. In this setting, a model was trained jointly on pairs of languages to analyze how the inclusion of a high-resource language influences representation learning for a low-resource language, and conversely, whether low-resource data affects performance in a high-resource setting, as described in Subsection~\ref{sec:bi-results}. We conclude our analysis by evaluating our fine-tuned models cross-lingually across multiple languages to test whether performance is driven by content rather than language, and to investigate content generalization, as shown in Figure~\ref{fig:xl-transfer-results}.


\subsection{Retrieval Task and Evaluation Settings}
\label{sec:retrieval-task}
We evaluate \textbf{metadata-to-document retrieval} performance as initialized in Subsection~\ref{sec:rag}. For each legal act, the introductory metadata block serves as the query, while the remaining substantive text (with metadata removed) forms the retrieval target. A query is considered correct if its corresponding ground-truth document is retrieved within the top-$k$ results.
To assess retrieval performance under different corpus conditions, we consider two complementary evaluation settings. In \textbf{Full-dataset search}, each test query is evaluated against a collection containing all documents (training, validation, and test) in the relevant language(s). In contrast, \textbf{Test-only search} restricts retrieval to the subset of held-out test documents.

The size of the test set varies across languages. This variation arises because some languages were introduced into EU legislation at later stages, resulting in fewer available legal acts for earlier years, and because a small number of documents were excluded due to data corruption or incomplete text extraction. The exact number of test queries per language is reported in Appendix~\ref{sec:test_set_size}.

\subsection{Evaluation Metrics}
\label{sec:eval-metrics}

Let $\mathcal{Q}$ be the set of test queries, and let $\mathrm{rank}(q)$ denote the rank position of the ground-truth document returned for query $q$ (with $\mathrm{rank}(q)=\infty$ if not retrieved).
We compute Top-$k$ accuracy as:
\begin{equation}
\mathrm{Acc@}k \;=\; \frac{1}{|\mathcal{Q}|}\sum_{q \in \mathcal{Q}} \mathbb{I}\big[\mathrm{rank}(q) \le k\big]
\end{equation}

where $\mathbb{I}[\cdot]$ is the indicator function. We report $\mathrm{Acc@1}$, $\mathrm{Acc@3}$, and $\mathrm{Acc@5}$.

\subsection{Monolingual Fine-Tuning}
\label{sec:mono-results}

In the monolingual setting, each model is fine-tuned on a dedicated language and evaluated on retrieval in that same language.
We chose three high-resource languages (EN, DE, FR) and two low-resource languages from the dataset corpus to test our hypotheses. 
Figure~\ref{fig:test-ds-test-search} summarizes Top-$k$ retrieval accuracy across all five languages for test queries evaluated against the test document collection, highlighting the impact of fine-tuning on retrieval quality. Across all evaluated languages, fine-tuning consistently improves retrieval performance compared to the corresponding pre-trained models, with gains observable at Top-1, Top-3, and Top-5. 
While high-resource languages showed consistently better performance even on the baseline, gains were observed across all languages. 
On the other hand, for low-resource languages, baseline performance was much lower, but fine-tuning brought it to levels comparable to those of high-resource languages.
While absolute accuracy varies by language and backbone, the effect direction is consistent: monolingual contrastive adaptation yields a more reliable ranking of the correct legislative act among the top retrieved results. 
This indicates that fine-tuning effectively aligns short metadata-style queries with their associated legal texts and that this benefit generalizes across multiple European languages.
We also report the results for all other languages in Appendix~\ref{apx:retrieval-results}.

\subsection{Bilingual Fine-Tuning}
\label{sec:bi-results}
We evaluate bilingual fine-tuning by training on a high-resource language (English) jointly with a low-resource language (Latvian), treating aligned versions of the same legal act as positives. 
Table~\ref{tab:conversion_performance} shows the results for the three models, showing baseline results, trained on English-only, Latvian-only, and EN\_LV together.

The results are mixed across models. For \texttt{E5}, fine-tuning across multiple languages has an additive effect, and retrieval performance improves when using both languages. This result is not consistent with the \texttt{Qwen} models. We find that, for both models, training on the dedicated language yields better results than training on both languages together. The performance using both languages for fine-tuning is, in most cases, better than without training at all, but shows no additive effect. 

In addition to these results, we find that bilingual fine-tuning does not improve retrieval performance on English compared with English-only training. Across all models, adding Latvian data neither enhances nor substantially degrades English Top-1 or Top-5 accuracy. This asymmetry suggests that bilingual training primarily benefits lower-resource languages by leveraging additional high-resource supervision, while preserving strong performance on high-resource languages without introducing negative transfer.

\begin{table}[t]
\centering
\small
\resizebox{\linewidth}{!}{%
\begin{tabular}{llcccc}
\hline
\textbf{Model} & \textbf{Train} & \multicolumn{2}{c}{\textbf{EN Eval}} & \multicolumn{2}{c}{\textbf{LV Eval}} \\
 &  & Top-1 & Top-5 & Top-1 & Top-5 \\
\hline
E5      & ORIG   & 81.06 & 95.59 & 72.91 & 90.10 \\
        & EN     & 89.43 & \textbf{97.80} & 82.29 & 94.27 \\
        & LV     & 87.22 & 96.03 & \textbf{84.37} & 97.39 \\
        & EN--LV & \textbf{90.30} & 97.35 & 83.85 & \textbf{97.91} \\
\hline
Qwen3-0.6B  & ORIG     & 85.02 & 97.36 & 60.41 & 86.45 \\
            & EN     & \textbf{94.27} & \textbf{99.12} & 74.47 & 92.18 \\
            & LV     & 91.18 & 98.67 & \textbf{82.82 }& 94.27 \\
            & EN--LV & 88.54 & 97.35 & 77.08 & \textbf{97.39} \\
\hline
Qwen3-4B    & ORIG     & 85.90 & 97.80 & 76.56 & 95.31 \\
            & EN     & \textbf{96.04} & \textbf{99.56} & 87.50 & \textbf{97.39} \\
            & LV     & 95.59 & 95.55 & \textbf{88.54} & 96.87 \\
            & EN--LV & 90.74 & 98.67 & 75.52 & 96.35 \\
\hline
\end{tabular}
}
\caption{Top-1 and Top-5 performance of three models trained on English (EN), Latvian (LV), and combined EN--LV data, evaluated on EN and LV datasets.}
\label{tab:conversion_performance}
\end{table}

\begin{figure*}[t]
  \centering
  \makebox[\textwidth][c]{%
    \includegraphics[width=1\textwidth]{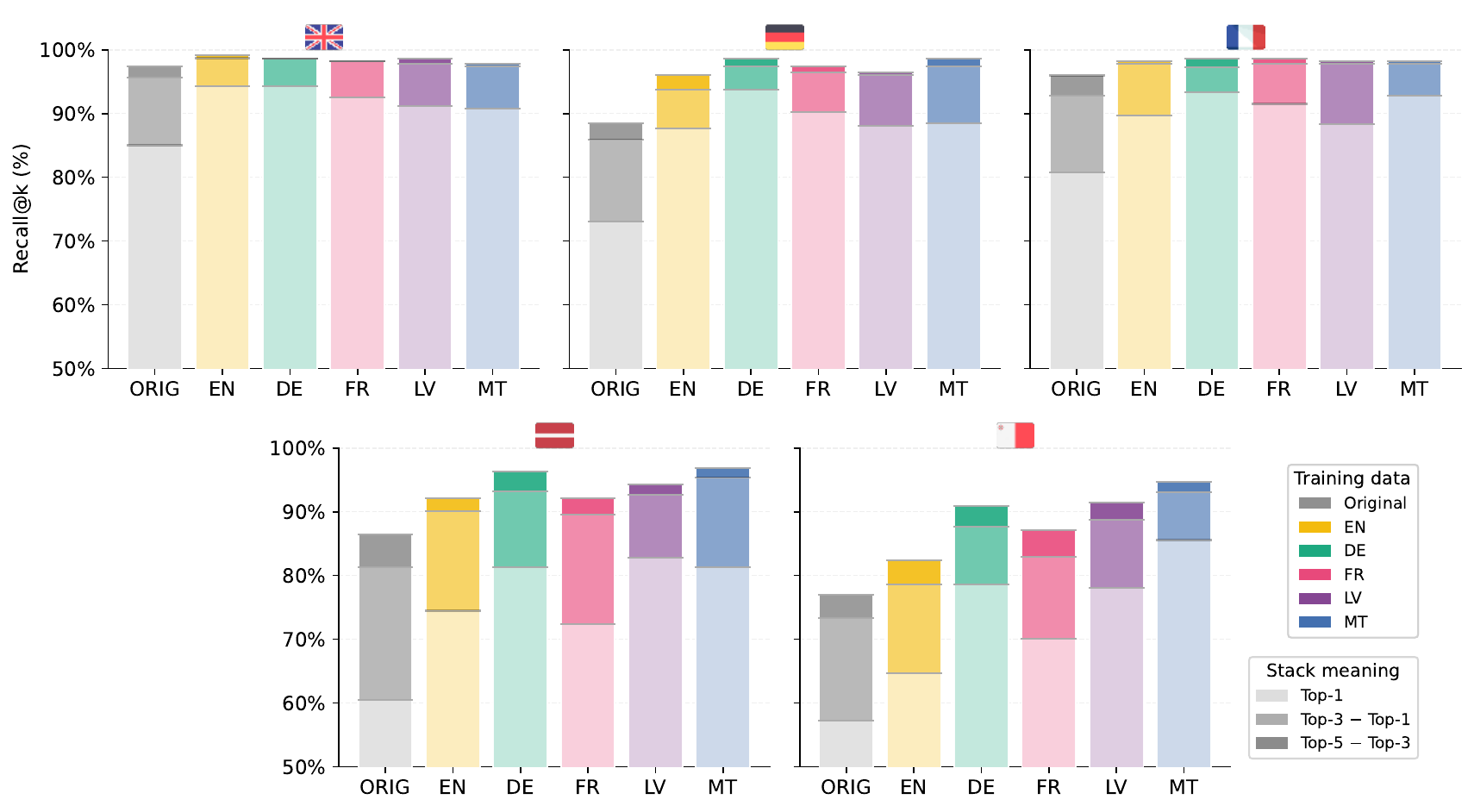}%
  }
  \caption{Cross-lingual fine-tuning for Qwen3 0.6B on five languages (EN, DE, FR, LV, MT). Performance is measured using Acc@\textit{k} for 1/3/5, with results presented as stacked bars, and compared between the base model and the fine-tuned variant.}
  \label{fig:xl-transfer-results}
\end{figure*}

\subsection{Cross-Lingual Transfer Results}
\label{sec:xl-transfer-results}
For further testing, regardless of whether the models learn language-independent general content, we evaluated the fine-tuned models on additional language evaluation datasets.
Therefore, models are fine-tuned on a source language and assessed on a different target language without further training. During evaluation, both queries and documents are in the target language, but embeddings are produced using the source-language fine-tuned model. This setting isolates the extent to which legal-domain knowledge learned in one language transfers to unseen languages. We conducted this experiment for each model and present the results in Figure~\ref{fig:xl-transfer-results} for \texttt{Qwen-0.6B} and for the others in Appendix~\ref{apx:cross_results}.

The results for the \texttt{Qwen3 0.6B} model again show differences between high- and low-resource languages. Across the high-resource languages (EN, DE, FR), the fine-tuned models generalize to other languages. For each language, we observe that Top-1 performance increases by at least 10\% relative to the baseline.
Top-5 performance is consistent across all three languages, with scores above 98\%, indicating that the task can be fully solved in other languages as well. 

For low-resource languages, we find that base performance is lower, but fine-tuning on other languages still yields higher results on those languages for Top-1 and Top-5. Improvements indicate that fine-tuning does not merely adapt the model to a specific language but instead enriches it with transferable legal-domain representations.

\subsection{Main Takeaways}

Across all experiments, fine-tuning embedding models on legal-domain data consistently improves metadata-to-document retrieval performance. Monolingual contrastive fine-tuning leads to higher Top-$k$ accuracy across languages and model sizes, indicating that domain-specific supervision helps models better capture the relationship between short metadata queries and their corresponding legislative texts.

Bilingual and cross-lingual evaluations further show that the improvements introduced by fine-tuning are not confined to the training language. Joint training with a high-resource language improves retrieval robustness for a lower-resource language. In contrast, cross-lingual evaluation shows that fine-tuned models generalize better than their original counterparts to unseen languages. Together, these observations suggest that fine-tuning primarily enhances content-level legal representations rather than relying on language-specific signals.

\section{Conclusion}
\label{sec:conclusion}


In this paper, we introduced \textsc{LEMUR}, a large-scale multilingual corpus of EU environmental legislation derived from official EUR-Lex PDFs. We proposed a unified framework for training and evaluating multilingual legal embedding models. To ensure data reliability, we introduced the Lexical Content Score (LCS), a systematic measure of PDF-to-text conversion quality. Using LEMUR, we fine-tuned three state-of-the-art embedding models on five  languages from the corpus. We evaluated them on metadata-to-document retrieval, reflecting realistic legal search scenarios.

Our results show that legal-domain contrastive fine-tuning consistently improves retrieval performance across languages and model sizes. Bilingual training further demonstrates that incorporating a high-resource language benefits retrieval in a low-resource setting without degrading high-resource performance. At the same time, cross-lingual evaluation confirms that these gains generalize beyond the training language. Together, these findings indicate that fine-tuning primarily enhances content-level legal representations rather than language-specific patterns. 

Future work will expand LEMUR to additional legal domains and languages and reduce the remaining PDF-to-text noise.

\section*{Limitations}
\paragraph{Limited topical coverage within EUR-Lex.}
LEMUR is restricted to \emph{Category~15} and \emph{Subcategory~10} (Environment). While this yields a focused benchmark, it limits topical diversity and may reduce generalizability to other EUR-Lex categories with different legal styles and terminology. Future work could extend collection and fine-tuning to additional categories and subcategories.
\paragraph{Limited bilingual fine-tuning coverage.}
Bilingual multi-positive fine-tuning is evaluated only on one language pair (EN--LV). Although this setting provides initial insights into bilingual training behavior, it does not explore the full range of possible language combinations available in LEMUR. Extending experiments to additional language pairs and resource configurations remains an important direction for future work.
\paragraph{Noise from PDF-to-text conversion.}
Although conversion quality is validated against HTML, the average lexical similarity across languages is about 0.94, indicating remaining extraction noise. Such noise can affect both fine-tuning and retrieval performance, particularly for older documents and lower-resource languages. Exploring alternative conversion pipelines and layout-aware post-processing could further improve text fidelity.

\section*{Acknowledgement}
This work is supported by the Genial4KMU project, Universität Hamburg, funded by BMBF (grant no. 16IS24044B).


\bibliography{clean2}

\appendix
\clearpage
\onecolumn

\section{Average Content Score for each language}
\label{sec:avg_content_score}

This table presents the average content score for each language.

\begin{table}[h]
\centering
\small
\begin{tabular}{l r}
\toprule
\textbf{Language} & \textbf{Avg. Content Score} \\
\midrule
English (EN)            & 0.9740 \\
Spanish (ES)            & 0.9734 \\
Dutch (NL)              & 0.9673 \\
Bulgarian (BG)          & 0.9671 \\
French (FR)             & 0.9608 \\
Romanian (RO)           & 0.9598 \\
Irish (GA)              & 0.9588 \\
Portuguese (PT)         & 0.9539 \\
Hungarian (HU)          & 0.9533 \\
Swedish (SV)            & 0.9520 \\
German (DE)             & 0.9487 \\
Italian (IT)            & 0.9463 \\
Croatian (HR)           & 0.9456 \\
Slovenian (SL)          & 0.9371 \\
Polish (PL)             & 0.9376 \\
Czech (CS)              & 0.9323 \\
Slovak (SK)             & 0.9294 \\
Greek (EL)              & 0.9135 \\
Latvian (LV)            & 0.9078 \\
Lithuanian (LT)         & 0.9067 \\
Finnish (FI)            & 0.9065 \\
Estonian (ET)           & 0.8991 \\
Maltese (MT)            & 0.8027 \\
\bottomrule
\end{tabular}
\caption{Average lexical content similarity between JSONL and HTML documents across languages.}
\label{tab:content_score_by_language}
\end{table}

\section{Content Score Comparison Across PDF-to-Text Conversion Methods}
\label{apx:content-score-methods}
\begin{figure*}[h]
  \centering
  \makebox[\textwidth][c]{%
    \includegraphics[width=\textwidth]{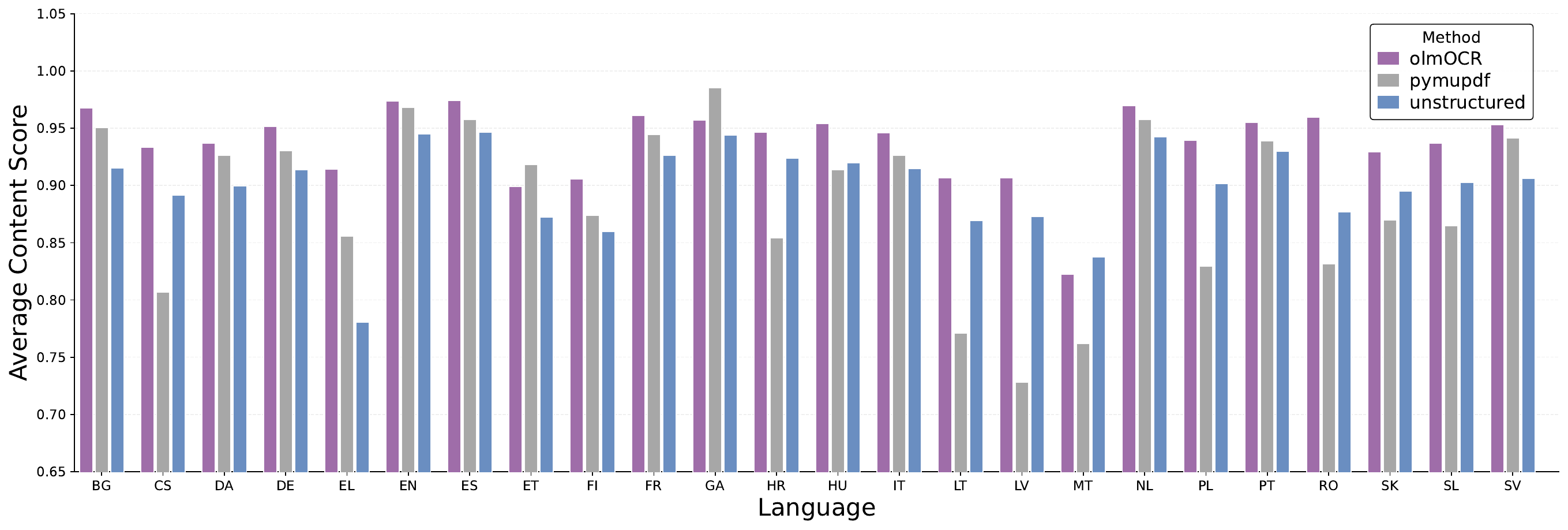}%
  }
\caption{Average Content Score per language for three PDF-to-text conversion methods. Scores are averaged over all documents available for each language.}
\label{fig:content-score-by-method}
\end{figure*}

This appendix provides a comparison of PDF-to-text conversion quality across languages and conversion pipelines. Figure~\ref{fig:content-score-by-method} reports the average Content Score for each language in LEMUR, computed separately for the three conversion methods used in our study: \textsc{olmOCR}, PyMuPDF, and Unstructured. Scores are averaged over all documents available for a given language and method.

\clearpage
\section{Content Score by Year and Dataset Coverage}
\label{apx:content-score-by-year}

\begin{figure*}[h]
  \centering
  \makebox[\textwidth][c]{%
    \includegraphics[width=\textwidth]{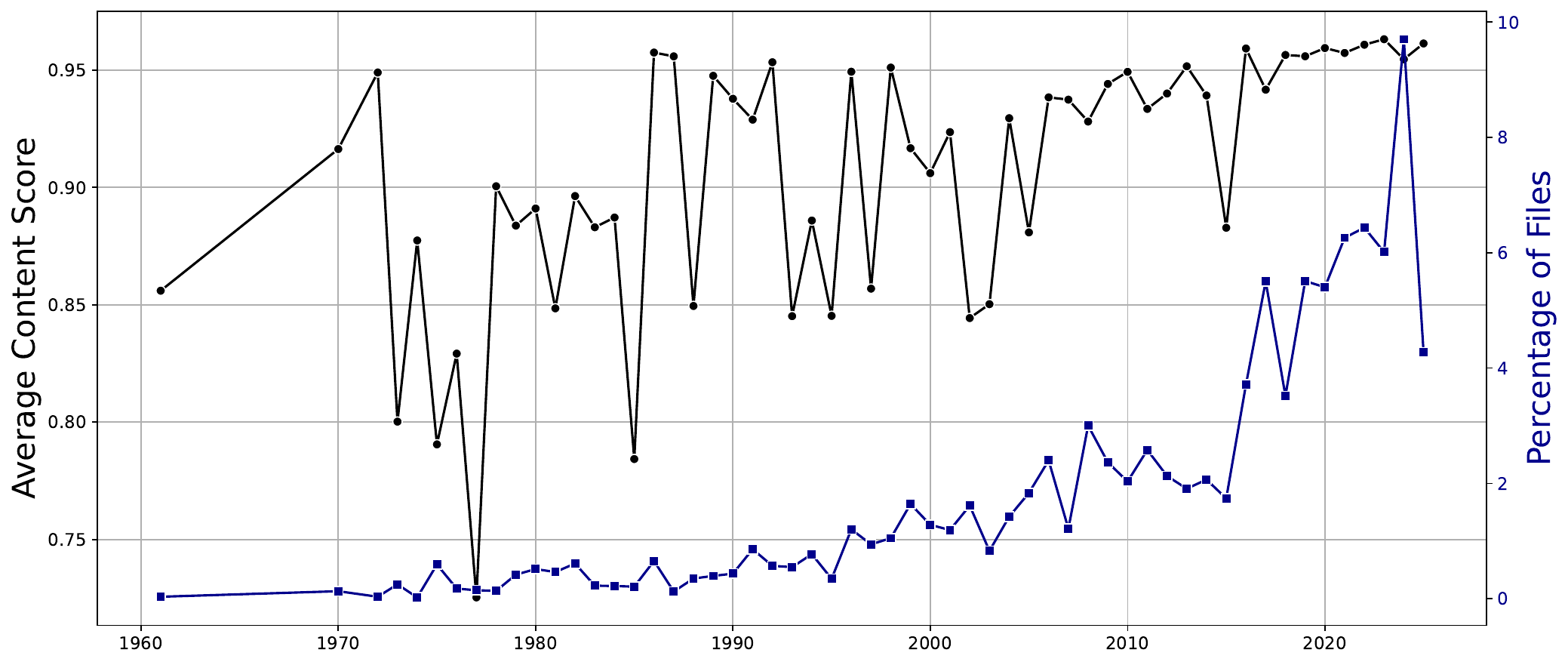}%
  }
  \caption{Average Content Score (left axis) and percentage of files (right axis) by publication year.}
  \label{fig:content-score-by-year}
\end{figure*}

This appendix reports how PDF-to-text conversion quality varies over time and how document availability is distributed across publication years. Figure~\ref{fig:content-score-by-year} plots (i) the average Content Score aggregated per year (left axis) and (ii) the corresponding percentage of files per year (right axis).

\clearpage
\section{Example Metadata--Document Pair}
\label{apx:dataset-examples}

\begin{figure*}[h]
  \centering
  \makebox[\textwidth][c]{%
    \includegraphics[width=0.8\textwidth]{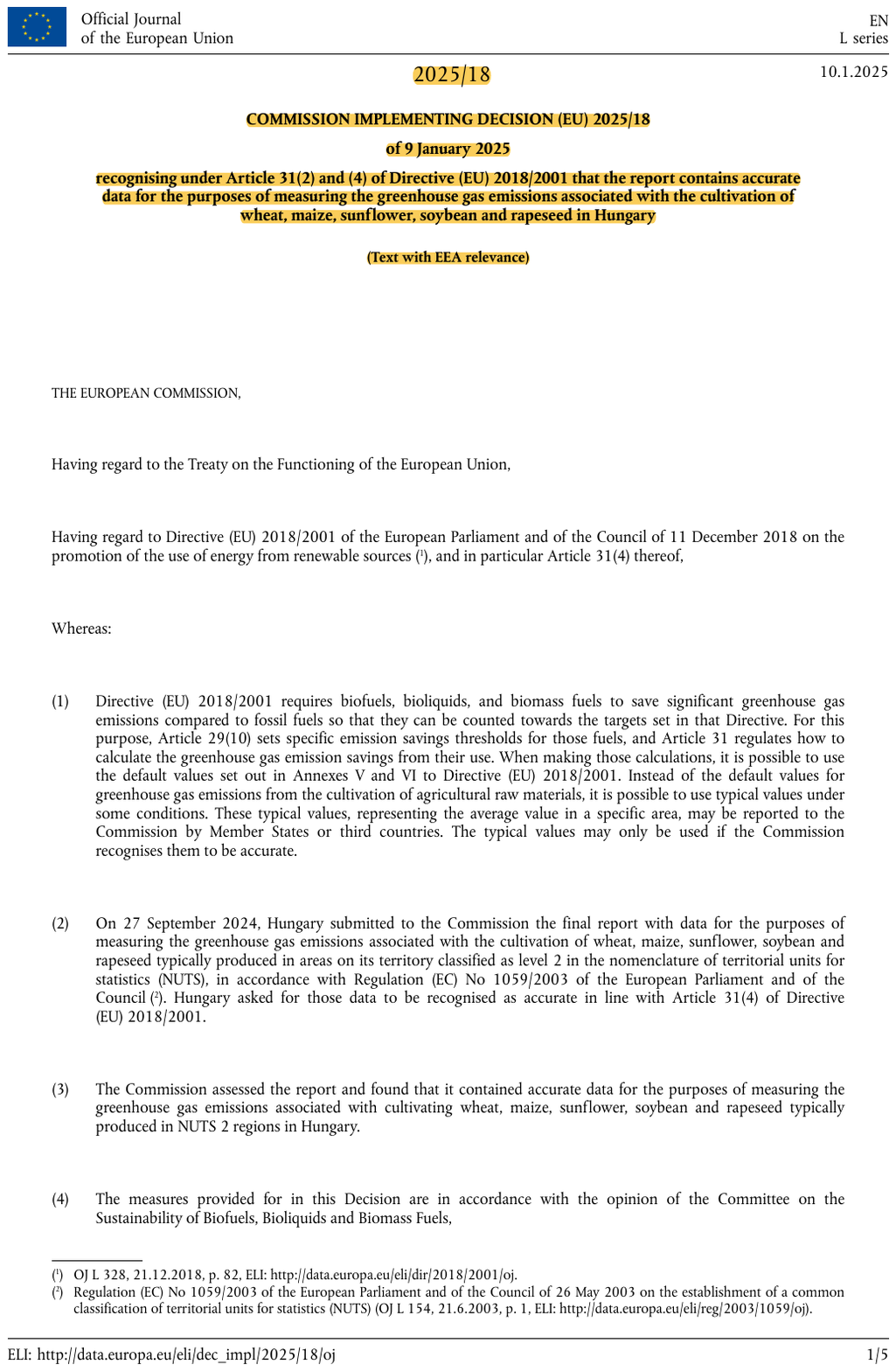}%
  }
  \caption{Example of a metadata--document pair in \textsc{LEMUR}.}
  \label{fig:metadata-document-example}
\end{figure*}

This appendix illustrates the metadata--document structure used throughout \textsc{LEMUR}.  
For each legislative act, the introductory metadata block is extracted and used as the retrieval query, while the remaining substantive legislative text constitutes the retrieval target.  
Figure~\ref{fig:metadata-document-example} shows a concrete example of this split for a single EU legislative document.

\clearpage

\section{Test Set Size for Each Language}
\label{sec:test_set_size}

This table reports the number of test queries available for each language used in the retrieval evaluation.

\begin{table}[h]
\centering
\small
\begin{tabular}{l r}
\toprule
\textbf{Language} & \textbf{Test Set Size} \\
\midrule
English (EN)  & 227 \\
German (DE)   & 226 \\
French (FR)   & 224 \\
Latvian (LV)  & 192 \\
Maltese (MT)  & 187 \\
\bottomrule
\end{tabular}
\caption{Number of test queries per language used in the retrieval evaluation.}
\label{tab:test_set_size_by_language}
\end{table}

\clearpage
\section{Cross-Lingual Results for E5 and Qwen-4B}
\label{apx:cross_results}

\begin{figure*}[h]
  \centering
\includegraphics[width=0.85\textwidth]{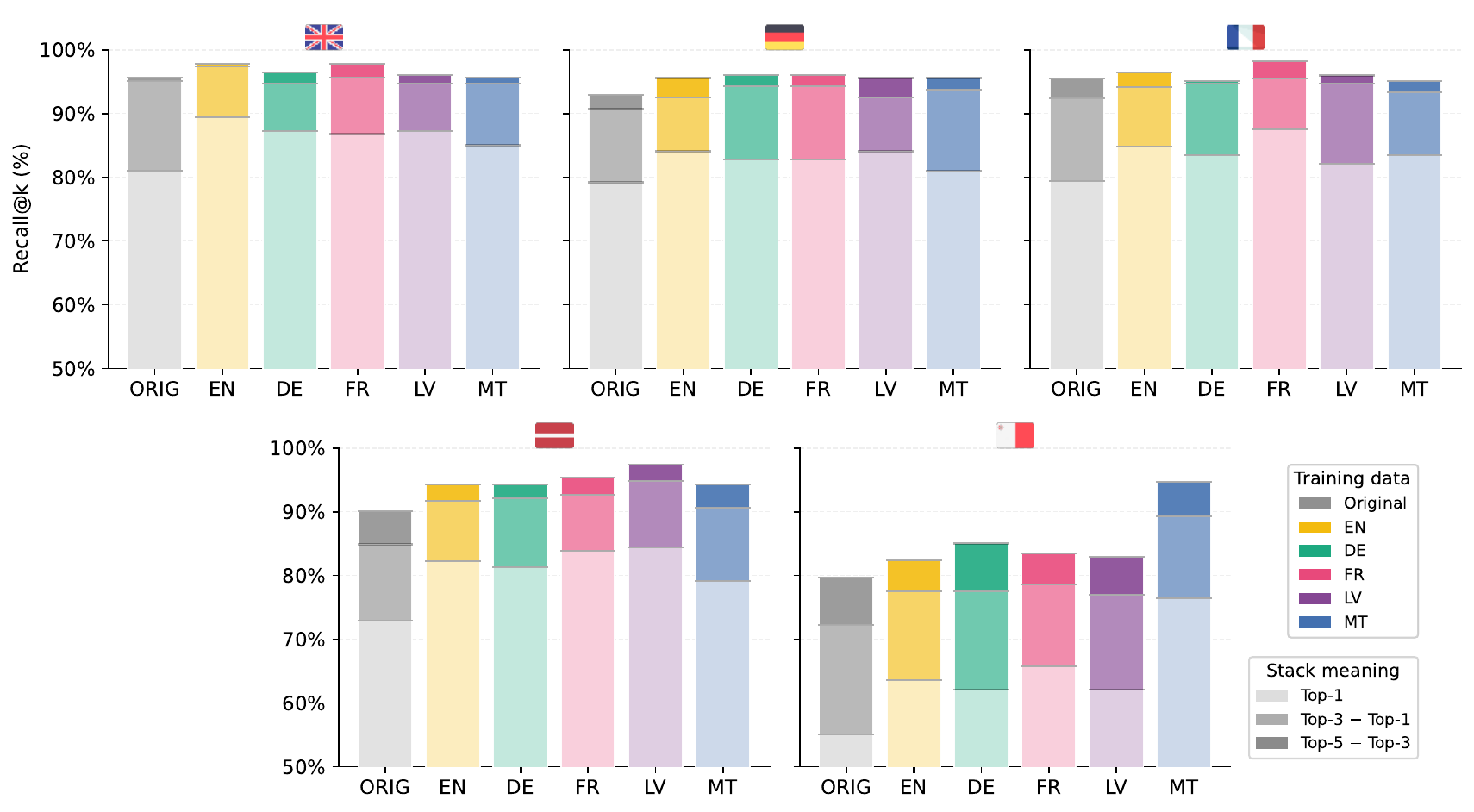}%
  \caption{Cross-lingual fine-tuning for E5 on five languages (EN, DE, FR, LV, MT). Performance is measured using Acc@\textit{k} for 1/3/5, with results presented as stacked bars, and compared between the base model and the fine-tuned variant.}
  \label{fig:e5-cross}
\end{figure*}

\begin{figure*}[h]
  \centering
  \makebox[\textwidth][c]{%
    \includegraphics[width=0.85\textwidth]{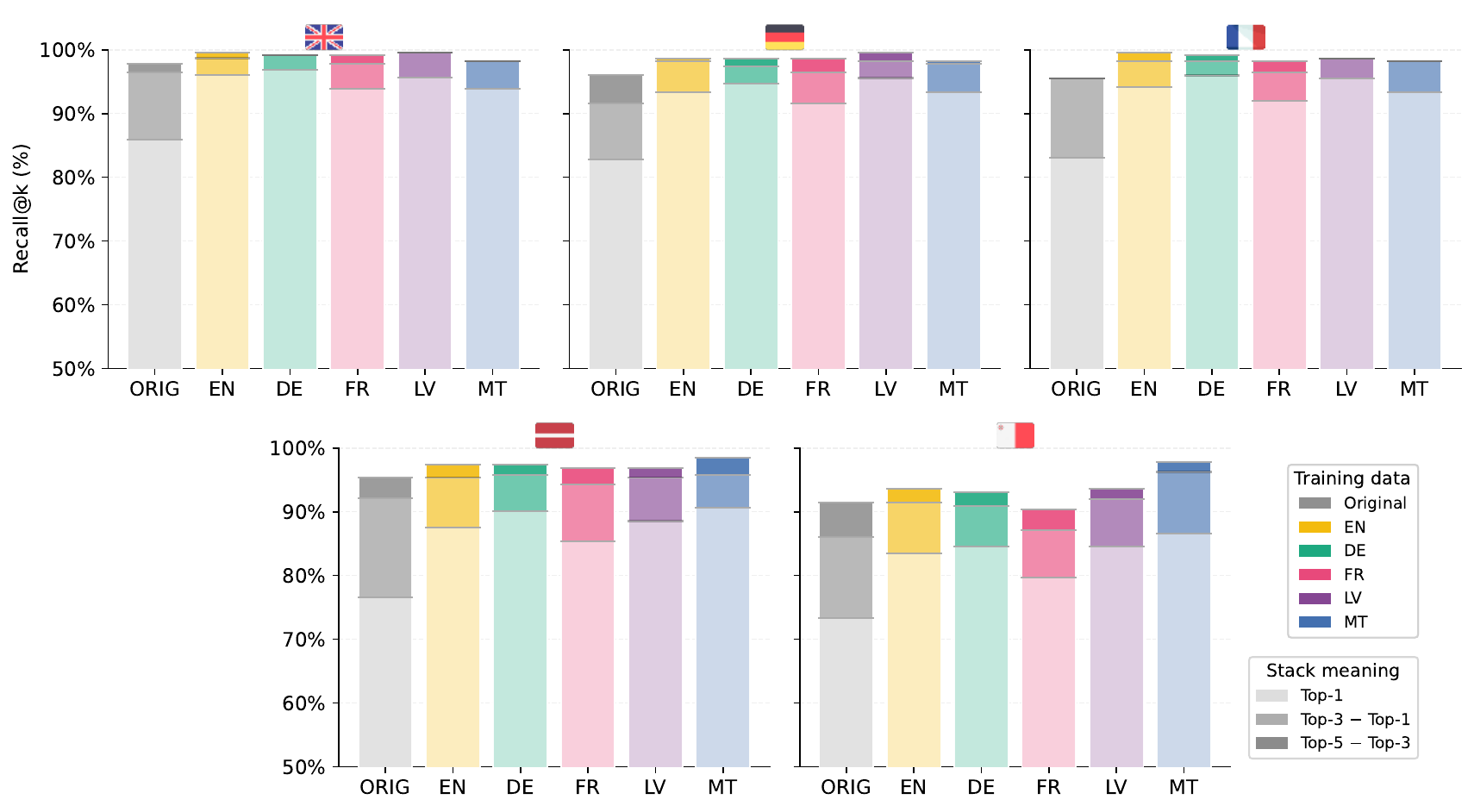}%
  }
  \caption{Cross-lingual fine-tuning for Qwen3-4B on five languages (EN, DE, FR, LV, MT). Performance is measured using Acc@\textit{k} for 1/3/5, with results presented as stacked bars, and compared between the base model and the fine-tuned variant.}
  \label{fig:qwen-4b-cross}
\end{figure*}

This appendix presents additional cross-lingual retrieval results for the \texttt{E5-Multilingual} and \texttt{Qwen3-4B} models. Figures~\ref{fig:e5-cross} and~\ref{fig:qwen-4b-cross} report Acc@\textit{k} ($k \in \{1,3,5\}$) for five target languages when models are fine-tuned on a single source language and evaluated cross-lingually without further adaptation.

\clearpage

\section{Monolingual Retrieval Performance with Test Queries over the Full Collection}
\label{sec:appendix-retrieval-full}

\begin{figure*}[h]
  \centering
  \makebox[\textwidth][c]{%
    \includegraphics[width=0.85\textwidth]{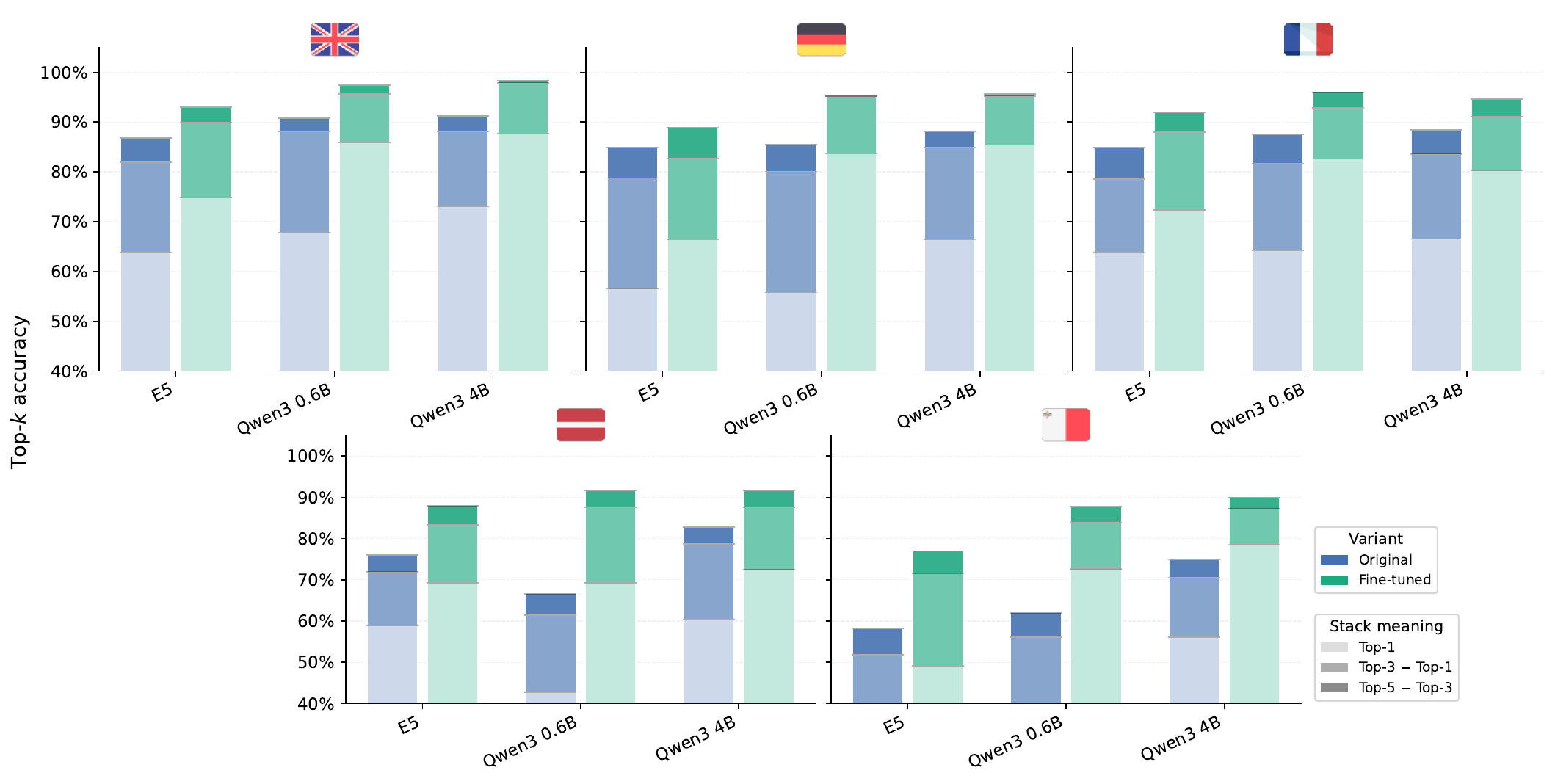}%
  }
\caption{Monolingual fine-tuning of three embedding models (E5, Qwen-0.6B \& Qwen-4B) on five languages (EN, DE, FR, LV, MT). Performance is measured using Acc@\textit{k} for 1/3/5 on test queries, evaluated against the full document collection, and is represented as stacked bars, with comparisons between the base model and the fine-tuned variant.}
\label{fig:full-ds-test-search}
\end{figure*}

Figure~\ref{fig:full-ds-test-search} shows monolingual retrieval performance when test queries are evaluated against the full document collection, including training, validation, and test documents. Results compare pretrained and fine-tuned models across five languages and three embedding backbones.

\clearpage
\section{Retrieval Results by Language}
\label{apx:retrieval-results}

This appendix reports monolingual retrieval results across eighteen languages.  
Table~\ref{tab:retrieval_test_docs} shows fine-tuned and original model performance for Acc@\textit{1} and Acc@\textit{5} when test queries are evaluated against \textbf{test documents only}, while Table~\ref{tab:retrieval_all_docs} shows results when test queries are evaluated against \textbf{all documents}.

\begin{table}[!tbh]
\centering
\begin{tabular}{l r r r r}
\toprule
\textbf{Language} & \textbf{Fine-tuned Top 1} & \textbf{Fine-tuned Top 5} & \textbf{Original Top 1} & \textbf{Original Top 5} \\
\midrule
Spanish (ES) & 83.03  & 96.87 & 77.23 & 92.18\\
Dutch (NL) & 84.95 & 96.01& 79.64& 95.57\\
Bulgarian (BG) & 80.31 & 96.27 & 73.40 & 88.29\\
Romanian (RO) & 86.55 & 98.38& 77.95&93.54 \\
Irish (GA) & 69.76 & 97.67& 39.53& 67.44\\
Portuguese (PT) & 81.61& 95.96& 77.57& 91.92\\
Hungarian (HU) &83.85 &93.75 &72.39 &90.62 \\
Swedish (SV) &89.20 & 99.53& 81.22& 95.30\\
Italian (IT) & 85.77 & 95.55 & 75.55 & 90.66\\
Croatian (HR) &82.08 &93.06 & 76.30& 86.70\\
Slovenian (SL) & 83.58&93.84 & 77.43& 91.28\\
Polish (PL) & 85.56 & 97.93& 77.83& 95.87\\
Czech (CS) & 86.08 & 96.90 & 78.35 & 94.84\\
Slovak (SK) & 85.64& 96.92& 77.43& 94.35\\
Greek (EL) & 77.00 & 95.00 & 74.00 & 90.00 \\
Lithuanian (LT) &82.98 & 93.81& 65.46&82.47 \\
Finnish (FI) & 80.46 & 93.02 & 69.76 & 87.44 \\
Estonian (ET) & 89.06 & 97.39& 78.12 & 93.75\\
\bottomrule
\end{tabular}
\caption{Retrieval results for test queries evaluated against test documents only.}
\label{tab:retrieval_test_docs}
\end{table}

\begin{table}[!tbh]
\centering
\begin{tabular}{l r r r r}
\toprule
\textbf{Language} & \textbf{Fine-tuned Top 1} & \textbf{Fine-tuned Top 5} & \textbf{Original Top 1} & \textbf{Original Top 5} \\
\midrule
Spanish (ES) &69.64  & 90.17 & 57.58& 81.69\\
Dutch (NL) & 66.37& 89.82& 55.30& 79.64\\
Bulgarian (BG) & 62.23 & 84.04 & 55.85 & 77.65 \\
Romanian (RO) &66.12 &89.78 &62.36 &85.48 \\
Irish (GA) & 34.88&76.74 &23.25 &48.83 \\
Portuguese (PT) & 65.47& 87.89& 54.26& 80.71\\
Hungarian (HU)  & 63.02 & 87.50 & 50.52 & 78.64\\
Swedish (SV) & 69.01 & 90.61& 61.03& 85.91\\
Italian (IT) & 71.55 & 89.77 & 56.88& 83.55\\
Croatian (HR) &63.00 &85.54 &57.22 & 78.03\\
Slovenian (SL) & 68.71 & 88.20&56.41 &77.94 \\
Polish (PL) & 69.58 & 89.69 & 58.76& 85.05\\
Czech (CS) & 72.68 & 90.72 & 61.34 & 85.05\\
Slovak (SK) &70.76 &87.17 &60.51 & 84.61\\
Greek (EL) & 60.00 & 85.50 & 58.00 & 79.00 \\
Lithuanian (LT) & 61.34 & 81.95& 50.00&75.77 \\
Finnish (FI) & 64.65 &85.11 & 53.48& 73.95\\
Estonian (ET) & 71.35 & 92.70& 60.41 & 85.41 \\
\bottomrule
\end{tabular}
\caption{Retrieval results for test queries evaluated against all documents.}
\label{tab:retrieval_all_docs}
\end{table}

\end{document}